\documentclass[conference]{IEEEtran}
\IEEEoverridecommandlockouts

\usepackage{cite}
\usepackage{amsmath,amssymb,amsfonts}
\usepackage{algorithmic}
\usepackage{graphicx}
\usepackage{textcomp}
\usepackage{url}
\usepackage{xcolor}
\usepackage{array}
\usepackage{multirow}
\def\BibTeX{{\rm B\kern-.05em{\sc i\kern-.025em b}\kern-.08em
    T\kern-.1667em\lower.7ex\hbox{E}\kern-.125emX}}
\begin{document}

\title{Structural Abstraction and Selective Refinement for Formal Verification
}

\author{\IEEEauthorblockN{1\textsuperscript{st} Christoph Luckeneder}
\IEEEauthorblockA{
Vienna, Austria \\
christoph.luckeneder@gmx.at
} 
\and
\IEEEauthorblockN{2\textsuperscript{nd} Ralph Hoch}
\IEEEauthorblockA{\textit{Institute of Computer Technology} \\
\textit{TU Wien}\\
Vienna, Austria \\
ralph.hoch@tuwien.ac.at}
\and
\IEEEauthorblockN{3\textsuperscript{rd} Hermann Kaindl}
\IEEEauthorblockA{\textit{Institute of Computer Technology} \\
\textit{TU Wien}\\
Vienna, Austria \\
hermann.kaindl@tuwien.ac.at}
}


\maketitle

\begin{abstract}

%

Safety
verification of robot applications is extremely challenging due to the complexity of the environment that a robot typically operates in.
Formal verification with model-checking provides guarantees but it may often take too long or even fail for complex models of the environment.
A usual solution approach is {\em abstraction}, more precisely {\em behavioral} abstraction.
Our new approach introduces {\em structural abstraction} instead, which we investigated in the context of voxel representation of the robot environment.
This kind of abstraction leads to {\em abstract voxels}.
We also propose a complete and automated verification workflow,
which is based on an already existing methodology for robot applications, and inspired by 
the key ideas behind counterexample-guided abstraction refinement (CEGAR) -- performing an initial abstraction and successively introducing refinements based on counterexamples, intertwined with model-checker runs.
Hence, our approach uses selective refinement of structural abstractions to improve the runtime efficiency of model-checking.
A fully-automated implementation of our approach showed its feasibility, since counterexamples have been found for a realistic scenario with a fairly high (maximal) resolution in a few minutes, while direct model-checker runs led to a crash after a couple of days.




\end{abstract}

\begin{IEEEkeywords}
structural abstraction, CEGAR, model checking, nuXmv
\end{IEEEkeywords}

\section{Introduction}

%
%

For verification through model-checking, the complexity of the formal models used is highly critical. A usual approach to address this is through abstraction, and we propose {\em structural abstraction} of the environment model.
Technically, our approach
approximates environment objects via a composition of cuboids of the same size called \emph{voxels}. 
The size of the voxels directly influences the accuracy of the approximation. Smaller voxels approximate objects better
than larger ones, where the latter are an abstraction of the former.
Since this abstraction does not involve any change in the {\em behavior} of the robot, we denote it as structural abstraction.

In this paper, we present an approach that reduces the verification time, see \cite{Luckeneder:2022}, which this paper is based upon.
It achieves this
by \emph{systematically abstracting} the environment model through voxels and by \emph{refining} them locally depending on verification results. This approach has the advantage that the models are coarse first and, hence, fast to verify, and only become more detailed in areas where verification runs of abstract models fail. To this end, our defined verification workflow starts with a representation using small voxels throughout, abstracts it by generating a representation consisting of larger voxels, still throughout, and from then on selectively adds the details needed for the actual verification where needed. 

The remainder of this paper is organized in the following manner.
First, we provide some background material in order to make this paper self-contained.
Then we present our new approach to selective refinement of structural abstraction.
For evaluating its feasibility, we present and explain the results of applying our new approach to verifying a safety-critical robot scenario.
Finally, we compare our approach with related work
and provide a conclusion.

\section{Background}

We provide some background material first on our running example of a robot arm performing a pick-and-place task. Subsequently, an existing methodology for verifying such robot applications is described. Since voxels and voxel grids play a major role in our paper, we give some background on them as well. Finally, we sketch \emph{counterexample-guided abstraction refinement} (CEGAR), a methodology to systematically abstract and refine behavioral models.

\subsection{Running Example}

For this paper, we reuse the running example presented in Rathmair~et~al.~\cite{rathmair2021}, where all the details are given. 
For the safety-critical aspects, see also below.

\begin{figure}[t]
\centerline{\includegraphics[width=1.0\columnwidth]{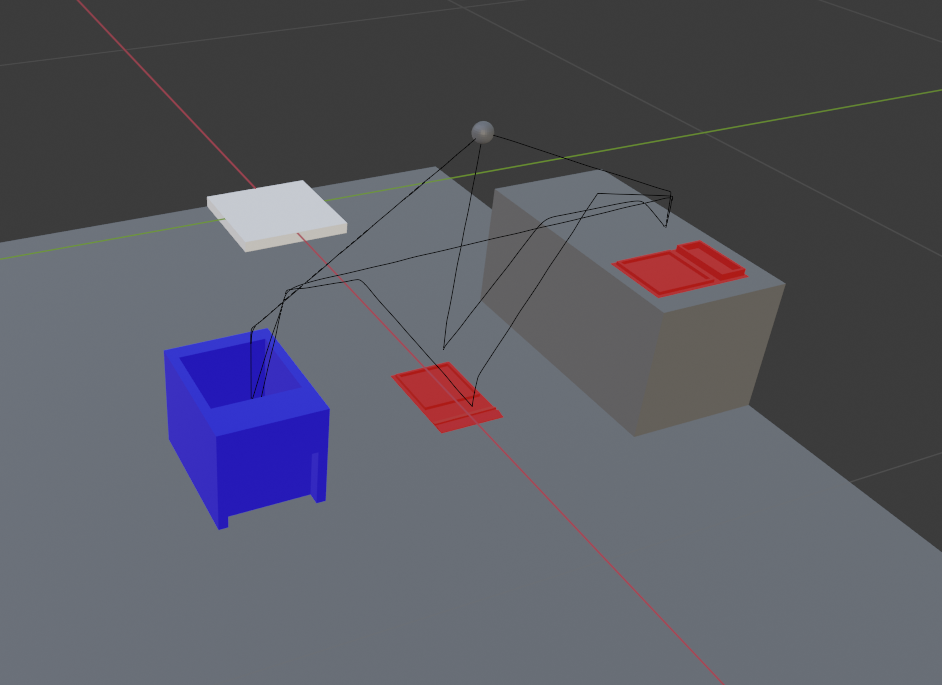}}
\vspace*{-2mm}

\caption{Environment model of the running example including gripper position (gray sphere) and trajectories (black lines)}
\label{fig:Background:RunningExample}
\end{figure}

In order to make this paper self-contained, we provide a short introduction to the use case of this running example here. 
Figure~\ref{fig:Background:RunningExample} illustrates a static environment where multiple pick-and-place tasks are to be performed, e.g., picking two objects at their initial position and placing them at different target locations.
A robot manipulator with a gripper performs these tasks and is 
mounted on the white base plate in the background of the figure.
At the start of the pick-and-place operation, a large object and a smaller one
are located on the red tray in the middle. The first task of the robot is to pick the large object and transfer it to the red tray located on the right. After placing the object on the tray, the robot moves back to the middle tray and picks a second smaller object, which is transferred to the blue box, where it is dropped into. After doing so, the robot arm moves back to the initial position, where it pauses.

The reason is that the robot performs this application in cooperation with a human, who manipulates the larger object while the robot moves the smaller object. After the human finishes his or her task, the robot picks the object from the tray, drops it into the blue box, and moves back to the initial position. This concludes a cycle of the application, and a new cycle may start.

We use this running example for verifying that no collision between the robot and its environment occurs. While verifying this scenario, we use a fixed trajectory with fixed positions of the robot. This is aligned with Rathmair~et~al.~\cite{rathmair2021}, where also the exact coordinates for the initial position and each position
on a trajectory 
from the initial to the end position
are used. Our proposed workflow operates on the environment representation only and, therefore, solely structural abstraction is performed. 



\subsection{Verification Methodology for Robot Applications}
\label{sec:Background:Methodology}

Rathmair~et~al.~\cite{rathmair2021} defined a generic verification approach for robot applications. 
It uses models describing the robot behavior and environment models and verifies the combined model against safety properties based on risk analyses, laws and standards. For details on this workflow, we refer to \cite{rathmair2021},
but let us briefly sketch it here in order to provide the context of our new model-checking approach.
The robot task is given as a behavioral tree, which defines
the execution sequence of robot skills. Together with the definition of these skills, the behavioral tree defines the behavior of the robot -- the behavioral model. This model is then transformed into a representation that the model-checker takes as its input. 

For verifying whether a robot application can be considered safe, a environment model is crucial. In the approach of  Rathmair~et~al.~\cite{rathmair2021}, it is given as a 3D-model of the relevant parts of the physical environment that the robot is operating in. This model is then transformed into a voxel grid, which is stored in the binvox file format~\cite{binvox}. A resolution of the voxel grid has to be chosen, which defines the number of voxels representing the environment and, in effect, its level of detail. Actually, this model is just an intermediate representation before a corresponding input for the model-checker tool is generated. 
The actual verification is done via the model-checker nuXmv~\cite{NuXMV, Cavada:2014}. It receives the behavioral model, the model of the environment, as well as safety properties as its input.

\subsection{Voxel Grid}

A \emph{voxel grid} is a construct used in 3D computer graphics, which represents a particular 3D space in terms of its properties. To accomplish this, the voxel grid is composed of individual voxels, which represent a value in the voxel grid, that are all of the same size. Instead of explicitly giving the position of the voxel in terms of 3D coordinates, the position is given relative to other voxels by indexing them. The 3D coordinates of a voxel are calculated using its $x$, $y$ and $z$-index, the 3D space the voxel grid covers, and the resolution of the voxel grid.

One way to build a voxel grid is to first define a cuboid (in 3D space) that it should represent. This space is then divided into smaller cuboids that are represented by voxels. The number of cuboids (voxels) is defined by the resolution of the voxel grid and is given via the resolution along each axis, e.g., $4 \times 8 \times 4$ divides the large cuboid into ($4 \times 8 \times 4 =$) $128$ smaller cuboids (voxels). Although different resolutions for each axis are possible, it is more common to use the same value for each axis, which is also a power of two, e.g., $4 \times 4 \times 4$ or $8 \times 8 \times 8$. Using this definition, each voxel of a grid with a certain resolution (e.g., $4 \times 4 \times 4$) can be seen as a composition of eight grid voxels with the subsequent higher resolution (e.g., $8 \times 8 \times 8$).

\begin{figure}[h]
\centerline{\includegraphics[width=1.0\columnwidth]{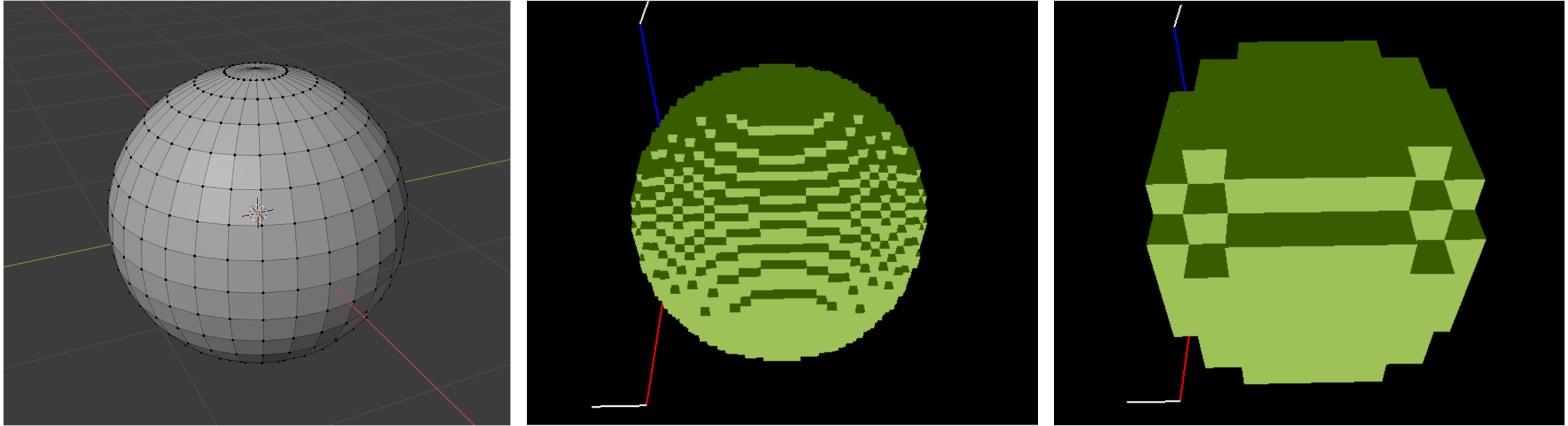}}
\vspace*{-2mm}

\caption{Voxel representation of a sphere in different resolutions -- left: sphere modeled in Blender~\cite{online:Blender}; middle: resolution $32 \times 32 \times 32$; right: resolution $8 \times 8 \times 8$}
\label{fig:Background:STL2Binvox:Sphere}
\end{figure}

In general, a higher resolution leads to a better approximation of the space, i.e., as more details can be represented, where a particular property holds or not. Figure~\ref{fig:Background:STL2Binvox:Sphere} shows the approximation of a sphere in different resolutions and indicates that higher resolutions are more precise. In this example, a voxel is filled in when its value is $True$, i.e., when the sphere at least partly occupies the voxel. Therefore, it results in an approximation that guarantees to enclose the whole sphere, independently of the resolution of the voxel grid. However, other forms of determining the value of the voxel are also reasonable, e.g., setting the voxel value to $True$ only if the entire voxel is part of the sphere. It entirely depends on the particular use case.

\begin{table*}[ht]
\begin{center}
	\caption{Results without Selective Refinement}
	\label{tab:results:WithoutRefinement}
	\vspace*{-2mm}
	\begin{tabular}{|l||*{7}{c|}}
	\hline
	Resolution 			& 2     & 4      & 8      & 16      & 32							& 64       		& 128 \\ \hline \hline
	Time         		& 0.1\,s & 0.2\,s & 0.4\,s & 12.2\,s & $\sim$19\,min 	& $\sim$50\,h & - \\ \hline
	Length					& 1     & 7      & 7      & 7       & 21    					& 80        	& - \\ \hline
	\end{tabular}%
\vspace*{2mm}\\
Resolution ... Resolution of Voxel Grid\\
Time ... Running Time of Verification\\
Length ... Length of Counterexample
\end{center}
\end{table*}

Usually, the voxel values are determined based on another form of representation of the 3D space. In our case, the sphere in the left part of Figure~\ref{fig:Background:STL2Binvox:Sphere} was modeled in Blender and exported to a file in the \emph{Standard Triangle Language (STL)} file format. We then used the tool \emph{binvox}~\cite{binvox} to generate the voxel grids based on the STL file, each with a different resolution. Finally, we used the tool \emph{viewvox}~\cite{viewvox} to view the generated voxel grids and capture the views given in Figure~\ref{fig:Background:STL2Binvox:Sphere}. However, the approach presented in this paper is not limited to these tools. For example, details of the 3D-space could be captured as a point cloud as well, which then would be used to generate a voxel representation.

\subsection{Counterexample-Guided Abstraction Refinement}
\label{background:CEGAR}

Counterexample-guided abstraction refinement (CEGAR)~\cite{clarke:1994, clarke2000, clarke:2003, clarke:2003_2, Clarke:Hybrid:2003} is an approach that uses a special form of abstraction 
(with one-sided abstraction error)
called \textit{over-approximation}, to reduce the state space in order to allow model checking of more complex systems. 
Intuitively,
an abstract model is an over-approximation of a concrete model, if it allows for all the behaviors of the latter and possibly more.
In the course of an abstraction, states of the concrete model are clustered into abstract states.
This may already lead to an increase of behavioral options through the transitions between clustered states in the abstract model.
However, no transition in the abstract model must be removed so that a possible behavioral option in the concrete model is not available in the abstract model.

Over-approximation guarantees that, if a temporal logic expression in \textit{ACTL*} \footnote{\textit{ACTL*} is a subset of \textit{CTL*} allowing only the path quantifier \textit{A}} evaluates to true in an abstract model, then it is true in the concrete model as well. If it evaluates to false in the abstract model, however, no conclusion can be drawn for the concrete model in this regard,
since it is not known whether the concrete or the abstract model caused the violation. Whenever the model violates the property, the model-checker tool generates a counterexample. 
The approach by
Clarke~et~al.~\cite{clarke:2003} uses it to first determine whether this counterexample is a valid path of the unabstracted model. If not, the information provided by the counterexample is used for refinement of the abstract model.

%


The CEGAR workflow as defined by Clark~et~al.~\cite{clarke:2003} starts with the computation of an initial abstraction. 
This is done by both taking the original model and the property to be verified into account. 
The abstract model is then checked against the given property, resulting either in a confirmation that the property holds, or a counterexample. In the former case the CEGAR verification workflow ends with \textit{pass}, in the latter it proceeds.   
If an abstract counterexample is generated, it is checked whether it is real or not (spurious). This is done by checking the counterexample on the unabstracted model and has two possible outcomes. If the counterexample is real (i.e., the abstract counterexample is realizable in the unabstracted model) the CEGAR workflow ends with \textit{fail}. If the counterexample is spurious (i.e., the counterexample is only present in the abstract model), the workflow refines the abstract model in such a way that the given counterexample is no longer admitted by the refined (abstract) model, when checked against the same property. 


%
%
%
%

In essence, CEGAR is an iterative workflow to verify \textit{ACTL*} properties by first generating a coarse abstraction of a given  model and gradually refining the abstraction based on spurious counterexamples until the given property holds in the abstract model (and, therefore, the unabstracted model).
Alternatively, it finds a counterexample that exists in the abstract model as well as in the unabstracted model.

\section{Selective Refinement of Structural Abstractions}

In this section, we present our new approach
to selective refinement of structural abstractions.
We start with motivating this approach for the structural abstraction of voxel grids used for environment representation. Then we present our approach for generating abstract voxels from more concrete ones. Based on it, we define our
verification workflow using structural abstraction and selective refinement. Finally, we explain how the workflow is integrated into the verification methodology of Rathmair~et~al.~\cite{rathmair2021},
where it can make the safety verification of certain robot applications 
much
more efficient.

\subsection{Why Structural Abstraction and Selective Refinements Matter}

Structural abstractions and their refinements matter because
the verification times of different voxel grid resolutions
are very different. 
Table~\ref{tab:results:WithoutRefinement} shows the running times needed to verify our running example (more precisely, the scenario leading to collision) with different resolutions, i.e., to find a counterexample in this case. The verification times range from a fraction of a second to $\sim$50\,h, depending 
only
on the 
different 
resolutions of the voxel grid.
For a resolution of $128$, it took the model-checker a couple of days to finally crash.

\begin{figure}[t]
\centerline{\includegraphics[width=1.0\columnwidth]{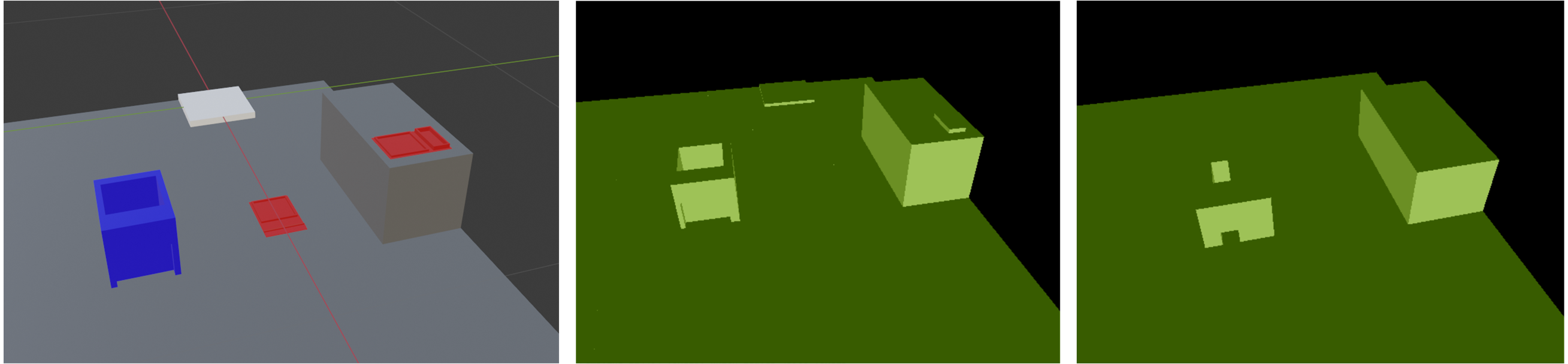}}
\vspace*{-2mm}

\caption{Voxel representation of 
environment model of the running example
--- left: modeled in Blender; middle: resolution $128 \times 128 \times 128$; right: resolution $32 \times 32 \times 32$}
\label{fig:StructuralAbstraction:STL2Binvox:Table}
\end{figure}


This strong variation in running times is due to the fact that 
the voxels that the robot can occupy in the 
model of the
environment increases
the more fine grained this model is, and this requires more calculations and comparisons with the voxels of the environment.
The verification times needed for each individual model-checker run increase strongly for higher resolutions, since the number of voxels increases with the power of $3$.
Not only do the verification times differ, but also the lengths of the counterexamples. For instance, the verification with a voxel grid of resolution $2$ ($= 2 \times 2 \times 2$) finds a counterexample of length $1$ (the property checked is already violated in the initial state). With a resolution of $4$, a counterexample of length $7$ is found.


As shown in Figure~\ref{fig:StructuralAbstraction:STL2Binvox:Table}, for representing the environment of a robot application, using lower resolution  -- which means a shorter verification time --  comes with the drawback of losing details. Whereas the higher-resolution (middle) voxel grid captures the top opening of the blue box quite well, the lower-resolution voxel grid does not. 
However, such details may be important when it comes to determining if a collision occurs. For example, when using the voxel grid with resolution $2 \times 2 \times 2$, the model-checker already detects a collision with the robot in its initial position. When using a resolution of $4 \times 4 \times 4$, no collision is detected at the same position. 
This effect of detecting a collision with a specific resolution that disappears with a higher resolution, is also the reason for the different counterexample lengths in Table~\ref{tab:results:WithoutRefinement}.

Unfortunately, one does not know upfront which details of the environment matter and, therefore, which resolution is needed to verify a property. Verification engineers can pick a resolution based on their experience or perform a detailed analysis of the whole robot application. In general, a higher resolution is preferable, since it means a more realistic representation of the environment than a lower resolution.
However, if the resolution is higher than needed, the model-checker run takes longer than necessary. If the resolution is too low, the verification may only fail due to the abstraction and the verification engineer has to determine if this is the case
or rerun the verification with a higher resolution.
A heuristic may be to use a resolution first that has verification runs in the order of a few minutes, and only increase it when necessary.


Using (automated) structural abstraction in combination with selective refinements 
as proposed below
allows mitigating or even overcoming those problems by using a voxel grid of low resolution first (automatically), thus reducing the time of verification, and {\em selectively} refining the voxel grid at points of particular interest to avoid finding counterexamples that only exist due to the low resolution used. 

\begin{figure}[t]
\centerline{\includegraphics[width=1.0\columnwidth]{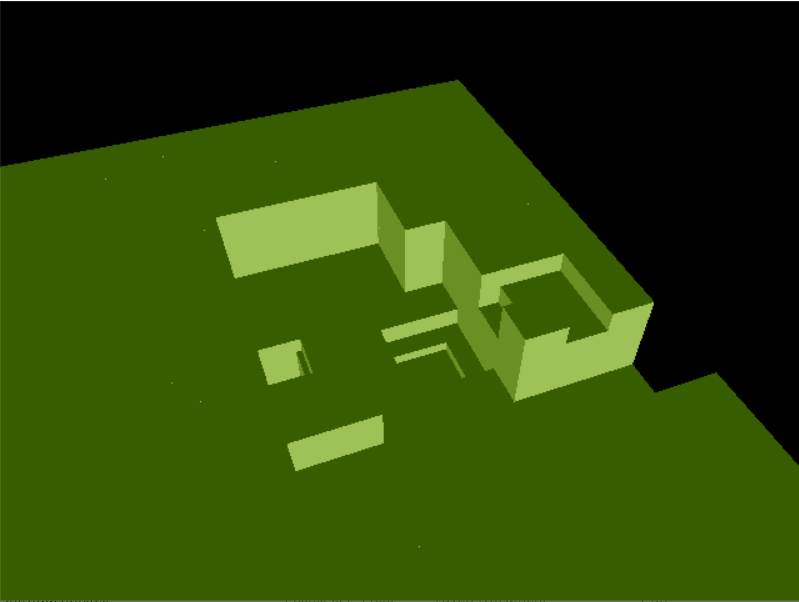}}
\vspace*{-2mm}

\caption{Environment representation with more details added by selective refinement}
\label{fig:StructruralAbstraction:Refinement}
\end{figure}

Figure~\ref{fig:StructruralAbstraction:Refinement} illustrates how selective refinement of a voxel grid with a 
given
resolution of 
$4 \times 4 \times 4$ may be changed to capture more details.
The resulting representation captures more details for certain parts of the environment, e.g., the opening of the box.

However, there is the challenge to determine where to refine the environment representation. This problem is specifically addressed by our new approach.

\subsection{
Abstracting Voxels
for Guaranteeing Over-approximation}
\label{section:AbstractionVoxels}

As stated above,
it is important how the abstraction is performed to guarantee that the abstract model is an over-approximation of the concrete model. 
Two or more voxels can always be joined together into one abstract voxel. Thus, there is always an abstraction function but it needs to be an over-approximation.
In this section, we show how voxels can be abstracted to lower the resolution of the voxel grid in such a way that guarantees over-approximation.  

CEGAR defines over-approximation for behavioral models only and not for structural ones. Therefore, we must dig deeper into what over-approximation does in the context of a particular property. Over-approximation allows the same or more behavior, and, consequently, the model-checker is more likely to encounter states where the atomic prepositions $x$ of a property like $\varphi = AG(x)$ becomes false. Therefore, it is also more likely that the entire property becomes false, which is detected by the model-checker.

Assuming that a voxel can be either (partly) occupied by an obstacle (SOLID) or free of an obstacle (not SOLID), a possible atomic proposition to check whether a collision occurs in a certain step is: 
\begin{itemize}
	\item $\alpha$ = voxels visited by the robot are \textit{not} SOLID
\end{itemize}
To check if the entire robot application is collision-free, the corresponding \textit{ACTL*} property is $\varphi = AG(\alpha)$, i.e., that the atomic proposition evaluates to true in each step.

Considering that the approximation does not alter the behavioral part of the model (the robot still moves along the same trajectory given in numerical coordinates), using different (structural) abstractions means checking which parts of the 3D space are visited by the robot in a different granularity (resolution). For example, this means checking $4 * 4 * 4 = 64$ voxels instead of  $8 * 8 * 8 = 512$ voxels if they are visited and SOLID.
However, to guarantee that all violations (collisions) detected by the finer-grained analysis are also detected by the coarser-grained one, the voxel values must be set right. 

Assuming we have two voxels, a free one and an occupied one named $v_0$ and $v_1$, respectively, $v_0$ has the value False (since it is not SOLID), and $v_1$ has the value True (since it is SOLID).
For reducing the number of voxels by joining them together into one abstract voxel $v_a$, it must be defined
which value is assigned to that voxel.
The abstract voxel $v_a$ needs to have the value True, in order to fulfill
over-approximation.

Considering an abstract voxel as SOLID when at least one (less abstract) voxel it is composed of is SOLID   
also corresponds nicely with the output of the tool \textit{binvox}. Given the same STL-file, \textit{binvox} generates voxel grids in such a way that an obstacle occupies more space in a lower-resolution voxel grid than in a higher-resolution one. Actually, we used voxel grids of different resolutions exported by \textit{binvox} to check this approach and to test the implementation of the abstraction mechanism.

In general, however, the abstraction approach depends on the property to be checked. Consider the following property:
\begin{itemize}
	\item $AG(voxels\,visited\,by\,the\,robot\,are\,FREE)$
\end{itemize}
Assume that a value of True means that the voxel is FREE.
Then the abstraction mechanism of 
setting the value of an abstract voxel to True when at least one of the more concrete voxels has a value of True does {\em not}
fulfill over-approximation.

\subsection{Our Verification Approach using Selective Refinement of Structural Abstractions}
\label{section:VerificationWorkflow}

The verification approach presented here uses structural abstraction of voxel grids in combination with selective refinements of individual voxels. It is inspired by the ideas behind CEGAR as we also perform an initial abstraction and improve the abstract model in incremental steps using information gathered by analyzing counterexamples.
In contrast to CEGAR, however, our approach works on structural abstractions rather than behavioral ones. 

\begin{figure}[h]
\centerline{\includegraphics[width=1.0\columnwidth]{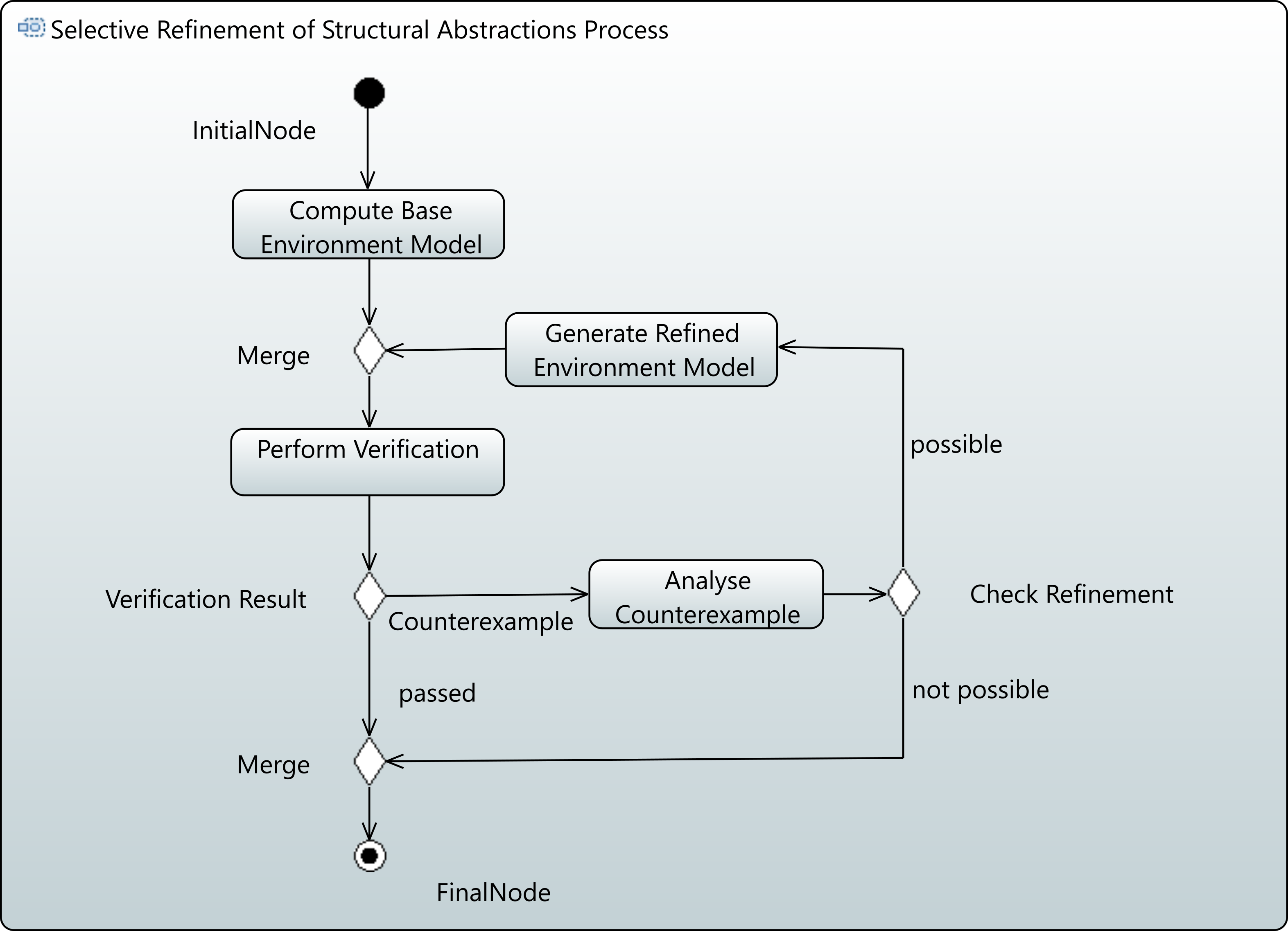}}
\vspace*{-2mm}

\caption{Verification workflow with selective refinement of structural abstractions}
\label{fig:StructuralAbstraction:Workflow}
\end{figure}

Our workflow, as illustrated in Figure~\ref{fig:StructuralAbstraction:Workflow}, consists of several steps. First, there is an initial step of abstracting the voxel grid provided to the workflow. After that, the whole workflow mainly operates on a voxel grid with reduced (as compared to the provided one) resolution. While the provided higher-resolution voxel grid is still available during the whole workflow, it is not directly used for verification.

After the initial abstraction is available, the verification loop starts. It consists of performing a verification run, analyzing the counterexample and correspondingly refining the environment model. Details on these steps are given below. 

The workflow ends if either the verification run does not detect a violation of the property (and, therefore, does not generate a counterexample) or the proposed refinement is considered not valuable or even impossible. In the first case, the workflow ends with the result that the verification has passed. In the second case, the verification fails and the workflow provides the counterexample that caused it. 

During the execution of the workflow, we distinguish between three resolutions of a voxel (grid). Those are:
\begin{itemize}
\item {\em Max-resolution:} The highest resolution used during the execution of the workflow. This resolution of the voxel grid is stored in a \textit{binvox}-file.
\item {\em Base-resolution:} This resolution is chosen by the verification engineer and is the resolution the voxel grid is reduced to at the start of the workflow. 
It represents the coarsest resolution that is used during the execution of the workflow. It is the resolution of the initial abstraction.
\item {\em Voxel-resolution:} This is the specific resolution of a particular voxel and is neither greater than Max-resolution nor smaller than Base-resolution. 
\end{itemize}

To understand the difference between these resolutions better, we give a short example. The voxel grid provided to the workflow has a resolution of $128$ ($128 \times 128 \times 128$) and, therefore, divides the space covered by the voxel grid into $2,097,152$ individual voxels. Hence, the Max-resolution of the workflow is $128$. 
The initial abstraction is set to generate a voxel grid with a resolution of $4$ ($4 \times 4 \times 4$), i.e., the Base-resolution of the workflow is $4$. Note, with its $64$ voxels, this voxel grid covers the same space as the $2,097,152$ voxels provided to the workflow. 
Therefore, those $64$ voxels have $32,768$ times the size of the voxels provided to the workflow.   
However, instead of explicitly spelling out the size of voxels, we just say that a voxel has a particular Voxel-resolution. In our example, all voxels in the voxel grid provided to the workflow have a Voxel-resolution of $128$ and all voxels of the voxel grid generated by the initial abstraction have a Voxel-resolution of $4$. After introducing refinements, the environment representation does have voxels of various sizes and, therefore, various Voxel-resolutions.

 

As already stated, the first step of our workflow is to generate a voxel grid with reduced resolution. Depending on the Max-resolution, the preset Base-resolution and the property to be checked, an abstraction according to our approach for abstracting voxels 
is done. In our example, this step determines the value associated with a voxel of resolution $4$ by joining $32,768$ voxels of resolution $128$ together.
All the $128$-resolution voxels a $4$-resolution voxel is composed of are joined. For example, the $4$-resolution voxel with index $x_4=0$, $y_4=1$, $z_4=0$ is composed of all $128$-resolution voxels whose index fulfills the following expressions $x_{128} \in [0, 31]$,  $y_{128} \in [32, 63]$ and $z_{128} \in [0, 31]$.

The first step in the verification loop executes \emph{Perform Verification}. During this step, all the models required (like environment model, behavioral model, etc.) are put together to form the overall model. This model is then handed over to the model-checker to be verified against the given property. 
While all the technical details can be found at \url{https://zenodo.org/record/7622703}, let us briefly sketch how voxel grids are represented for passing them into the model-checker, both the unrefined and the abstracted ones.
There is a set of arrays, where one array stores the largest voxels with the highest resolution, and for each refinement of a voxel there is an extra array. An index structure serves for using the right arrays to be encoded into the input language of the model-checker.

The output of the model-checker is the result of this step and is either the statement that the property is verified successfully, or a counterexample, which shows that the property is violated.
Depending on this output, the workflow either stops with the result of successful verification of the model against the property, or it proceeds with executing the next action \emph{Analyze Counterexample}.

\emph{Analyze Counterexample} analyzes the output of the previous verification step. In our implementation, the output is captured as a log file and, hence, the file is parsed and the (human-readable) counterexample documented in it is analyzed fully automatically. The counterexample contains information on how variables of the model change at each time step of the verification run. With this information and knowing the names of the variables used to represent the index of a voxel, the specific voxel that caused the property violation  and the time step when the violation occurred can be determined. Based on this information, the action then outputs a suggestion for voxels to be refined. In our implementation, the suggestion is the specific voxel causing the violation. However, as we explain in the discussion section below, a more elaborate suggestion containing multiple voxels may be possible. 

The suggested refinement can only be performed if the voxel does not have Max-resolution already. Otherwise, no further refinement is possible and the workflow ends with the result of a failed verification.

The actual refinement is performed in the \emph{Generate Refined Environment Model} action. In this step, the (abstract) voxel is refined into eight new voxels with the next higher resolution, e.g., a voxel of resolution $8$ is refined into eight voxels of resolution $16$. 
To determine the Max-Resolution voxels to be combined to form a particular new voxel, the position of the voxel to be refined and the position of the new voxel inside of the refined one are used.  
To compute the value (SOLID or not) for each new voxel, the identified Max-Resolution voxels are combined.

Finally, the whole environment -- consisting of the voxel grid in base resolution and the refined ones -- is exported to its nuXmv input representation.  

The newly generated environment representation is then used during the next \emph{Perform Verification} step of the workflow. The cycle of verification, analyzing a counterexample, and generating a new environment representation continues until the workflow terminates.



\subsection{Integration into Preexisting Methodology}


The proposed workflow is integrated into the preexisting methodology of Rathmair~et~al.~\cite{rathmair2021}
for its application to safety verification. 
Compared to the original methodology, only minor changes were necessary. 
Instead of verification directly using the model-checker nuXmv,
our tool implementing the new verification workflow
is used, which invokes nuXmv as specified above. In addition to that, adapting the \emph{Robot Environment Path} was also necessary, since the original methodology would hand over an SMV model of the environment. However, our workflow asks for binvox-files representing the environment. Hence, both processes of the original \emph{Robot Environment Path} were replaced by a new process generating the binvox-file(s) needed. As defined above, the resolution of this binvox-file is also the Max-resolution of the workflow. Since our workflow also uses the model-checker nuXmv, it has to generate the SMV models of the environment. 

\section{Results}




\begin{table*}[ht]
\begin{center}
\caption{Results with Selective Refinement -- Max-Resolution 128}
\label{tab:results:WithRefinement}
\vspace*{-2mm}
\begin{tabular}{|l||*{7}{c|}}
\hline
Base-Resolution 			& 2     & 4      & 8      & 16      & 32							& 64       \\ \hline \hline
Time         		& $\sim$4\,min & $\sim$3\,min & $\sim$3.5\,min & $\sim$14\,min & $\sim$6\,h 	& $\sim$8.4\,d \\ \hline
Length					& 80     & 80      & 80      & 80       & 80    					& 80       \\ \hline
Refinements 		& 39		& 35			& 32		& 23			& 13							& 3					\\ \hline

\end{tabular}%
\vspace*{2mm}\\
Base-Resolution ... Base-Resolution used in the Workflow\\
Time ... Running Time of Verification \\
Length ... Length of Counterexample\\
Refinements ... Number of Refinements made
\end{center}
\end{table*}




We applied our new approach to verifying 
our running example, which originated from Rathmair~et~al.~\cite{rathmair2021}
and is also sketched above.
Actually, we model-checked one cycle only.
Assuming that the cycles are identical, the verification of one cycle is sufficient. 
Table~\ref{tab:results:WithoutRefinement} above shows that even for a lower resolution of 64, $\sim$50\,h were needed for finding a counterexample.
These results correspond to the performance of the approach without selective refinement by Rathmair~et~al.~\cite{rathmair2021}.
While the amount of memory needed increases with the more fine-grained environment models, it was never a problem in our model-checking runs.
Hence, we only report running times.
All model-checking runs in this paper were performed on an Ubuntu 20.04 system running on an AMD Ryzen 7 5800X 8-Core Processor with 3.8GHz, 64GB memory and a GeForce GT 710. 

Using these results from previous work for comparison,
we present and explain the results
of applying our new approach 
for demonstrating that it can strongly improve model-checking performance.
More precisely,
we verified this scenario using our workflow with a Max-resolution of $128$. Note again, that there were no results for verification without selective refinement for this high resolution, since the model-checker did not finish with a result even after a couple of days.

Table~\ref{tab:results:WithRefinement} shows the results for our approach with selective refinement, for various values of Base-resolution, which has to be set before each verification run as a kind of parameter. A Base-resolution of $128$ does not make sense for a Max-resolution with the same value, since using a Base-resolution equal to the Max-resolution leads to the same model-checker run as without selective refinement. Hence, there is no related result given in this table.
Also the verification with a Base-resolution of $64$ took extremely long, but using such a high Base-resolution is not reasonable since it
constrains our approach too much. 

The lowest running times in the order of a few minutes have been achieved with low values for Base-resolution, i.e., when starting the model-checker runs with low resolutions, which finish fairly quickly. After that, selective refinement shows its benefits by exploring higher resolutions only where necessary for finding a real counterexample.
Hence, with lower values of Base-resolution, a (real) counterexample for Max-resolution of 128 was possible to be found within a few minutes, while the model-checker directly running with the high resolution and without selective refinement did not finish with a result even after a couple of days.

\begin{figure}[h]
\centerline{\includegraphics[width=1.0\columnwidth]{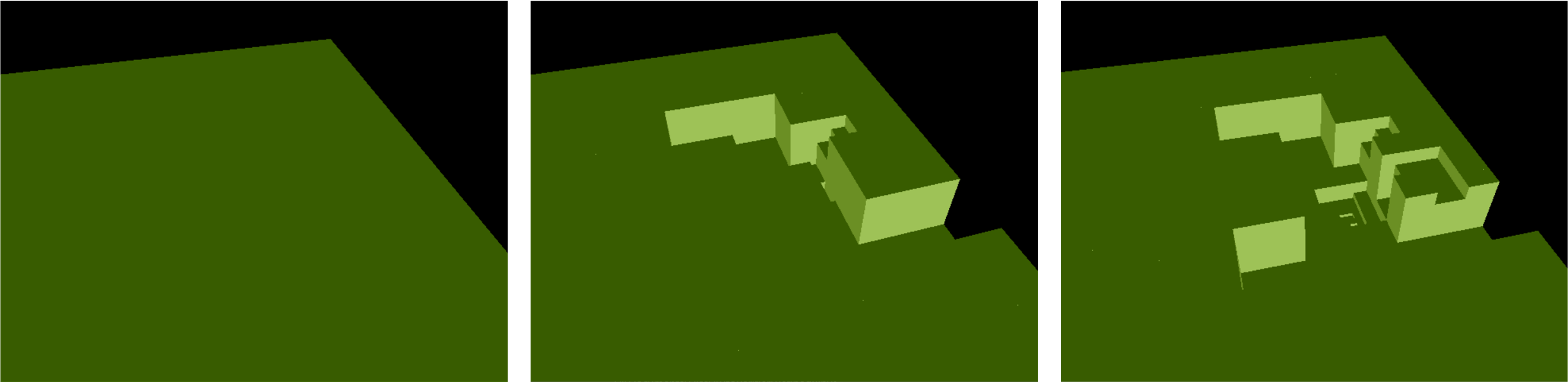}}
\vspace*{-2mm}

\caption{Evolution of the voxel grid with Base-resolution $4$ and Max-resolution $128$ -- left: without any refinement; middle: with $16$ refinements; right: with $35$ refinements}
\label{fig:Results:VoxelGrid}
\end{figure}

For the purpose of illustrating selective refinements,
Figure~\ref{fig:Results:VoxelGrid} visualizes an example of the evolution of the voxel grid when applying the workflow for performing refinements with a Base-resolution of $4$ and Max-resolution of $128$. The leftmost subfigure shows the environment in a resolution of $4$. Neither the blue box nor the raised tray are visible. After a few refinements, both can already be seen vaguely. With its $35$ refinements, the final representation gives even more details. Note, that the opening of the box seems not essential to disprove the property (in this case).

\begin{footnotesize}
\begin{table*}[ht]
\begin{center}
\caption{Verification-Times of all Base- and Max-Resolution Combinations for the \textit{Collision} Scenario}
\label{tab:results:Matrix}
\vspace*{-2mm}
\begin{tabular}{l c||*{6}{c|}}
\cline{3-8}		
\multirow{2}{*}{}
								& & \multicolumn{6}{c|}{Base-Resolution} \\ \cline{3-8}		
								& & 2     & 4      & 8      & 16      & 32							& 64       \\ \cline{3-8} \hline \hline
\multicolumn{1}{|l|}{\multirow{7}{*}{\rotatebox[origin=c]{90}{Max-Resolution}}}
								&\multicolumn{1}{|c||}{2} & 0.2\,s & - & - & - & - & - \\ \cline{2-8}
\multicolumn{1}{|l|}{}	& \multicolumn{1}{|c||}{4} 	& 0.6\,s & 0.4\,s  & -     & -     & -  	& -      \\ \cline{2-8}
\multicolumn{1}{|l|}{}  & \multicolumn{1}{|c||}{8} 	& 1\,s		 & 0.8\,s		& 0.6\,s		 & -		 & -		& -			\\ \cline{2-8}
\multicolumn{1}{|l|}{}  & \multicolumn{1}{|c||}{16} & 1.4\,s     & 1.2\,s      & 1.2\,s    & 12.4\,s   & -    & -       \\ \cline{2-8}
\multicolumn{1}{|l|}{}  & \multicolumn{1}{|c||}{32} & 7.2\,s     & 7\,s      & 8.5\,s      & $\sim$1\,min      & $\sim$19.4\,min & -   \\ \cline{2-8}
\multicolumn{1}{|l|}{}  & \multicolumn{1}{|c||}{64} & $\sim$2.7\,min & $\sim$2.5\,min & $\sim$2.6\, min & $\sim$12\,min & $\sim$5.3\,h   & $\sim$50\,h  \\ \cline{2-8}
\multicolumn{1}{|l|}{}  & \multicolumn{1}{|c||}{128} & $\sim$4\,min & $\sim$3\,min & $\sim$3.5\,min & $\sim$14\,min & $\sim$6\,h 	& $\sim$8.4\,d \\ \hline

\end{tabular}%
\vspace*{2mm}\\
Base-Resolution ... Base-Resolution used during the Workflow\\
Max-Resolution ... Max-Resolution used during the Workflow
\end{center}
\end{table*}

\begin{table*}[ht]
\begin{center}
\caption{Verification-Times of all Base- and Max-Resolution Combinations for the {\em Near Miss} Scenario}
\label{tab:results:Near}
\vspace*{-2mm}
\begin{tabular}{l c||*{6}{c|}}
\cline{3-8}		
\multirow{2}{*}{}
								& & \multicolumn{6}{c|}{Base-Resolution} \\ \cline{3-8}		
								& & 2     & 4      & 8      & 16      & 32							& 64       \\ \cline{3-8} \hline \hline
\multicolumn{1}{|l|}{\multirow{7}{*}{\rotatebox[origin=c]{90}{Max-Resolution}}}
								&\multicolumn{1}{|c||}{2} & 0.3\,s & - & - & - & - & - \\ \cline{2-8}
\multicolumn{1}{|l|}{}	& \multicolumn{1}{|c||}{4} 	& 0.8\,s & 0.5\,s  & -     & -     & -  	& -      \\ \cline{2-8}
\multicolumn{1}{|l|}{}  & \multicolumn{1}{|c||}{8} 	& 1.3\,s		 & 1.1\,s		& 0.8\,s		 & -		 & -		& -			\\ \cline{2-8}
\multicolumn{1}{|l|}{}  & \multicolumn{1}{|c||}{16} & 4.3\,s     & 5.3\,s      & 4\,s    & 18.3\,s   & -    & -       \\ \cline{2-8}
\multicolumn{1}{|l|}{}  & \multicolumn{1}{|c||}{32} & $\sim$1.4\,min     & $\sim$1.3\,min      & $\sim$1.4\,min      & $\sim$4\,min      & $\sim$35\,min & -   \\ \cline{2-8}
\multicolumn{1}{|l|}{}  & \multicolumn{1}{|c||}{64} & $\sim$6\,min & $\sim$5.5\,min & $\sim$5.6\,min & $\sim$17\,min & $\sim$7.3\,h   & $\sim$2.8\,d  \\ \cline{2-8}
\multicolumn{1}{|l|}{}  & \multicolumn{1}{|c||}{128} & $\sim$31.5\,min & $\sim$27.2\,min & $\sim$27.5\,min & $\sim$1.4\,h & $\sim$33.5\,h 	& $\sim$xxx\,d \\ \hline
\end{tabular}%
\vspace*{2mm}\\
Base-Resolution ... Base-Resolution used during the Workflow\\
Max-Resolution ... Max-Resolution used during the Workflow
\end{center}
\end{table*}

\begin{table*}[ht]
\begin{center}
\caption{Verification-Times of all Base- and Max-Resolution Combinations for the {\em Obviously Safe} Scenario}
\label{tab:results:Safe}
\vspace*{-2mm}
\begin{tabular}{l c||*{6}{c|}}
\cline{3-8}		
\multirow{2}{*}{}
								& & \multicolumn{6}{c|}{Base-Resolution} \\ \cline{3-8}		
								& & 2     & 4      & 8      & 16      & 32							& 64       \\ \cline{3-8} \hline \hline
\multicolumn{1}{|l|}{\multirow{7}{*}{\rotatebox[origin=c]{90}{Max-Resolution}}}
								&\multicolumn{1}{|c||}{2} & 0.3\,s & - & - & - & - & - \\ \cline{2-8}
\multicolumn{1}{|l|}{}	& \multicolumn{1}{|c||}{4} 	& 1\,s & 0.8\,s  & -     & -     & -  	& -      \\ \cline{2-8}
\multicolumn{1}{|l|}{}  & \multicolumn{1}{|c||}{8} 	& 1.9\,s		 & 2.4\,s		& 1.2\,s		 & -		 & -		& -			\\ \cline{2-8}
\multicolumn{1}{|l|}{}  & \multicolumn{1}{|c||}{16} & $\sim$1.3\,min & $\sim$1.2\,min  & $\sim$1\,min    & $\sim$1\,min   & -    & -       \\ \cline{2-8}
\multicolumn{1}{|l|}{}  & \multicolumn{1}{|c||}{32} & $\sim$1.7\,min     & $\sim$1.8\,min      & $\sim$1.3\,min      & $\sim$1.9\,min      & $\sim$41\,min & -   \\ \cline{2-8}
\multicolumn{1}{|l|}{}  & \multicolumn{1}{|c||}{64} & $\sim$3.3\,min & $\sim$3\,min & $\sim$2.6\,min & $\sim$5\,min & $\sim$1.8\,h   & $\sim$xxx\,d  \\ \cline{2-8}
\multicolumn{1}{|l|}{}  & \multicolumn{1}{|c||}{128} & $\sim$3.3\,min & $\sim$3\,min & $\sim$2.6\,min & $\sim$5\,min & $\sim$1.8\,h 	& $\sim$xxx\,d \\ \hline

\end{tabular}%
\vspace*{2mm}\\
Base-Resolution ... Base-Resolution used during the Workflow\\
Max-Resolution ... Max-Resolution used during the Workflow
\end{center}
\end{table*}

\end{footnotesize}

We also investigated the relative performance depending on the parameters Base- and Max-resolutions. For that, we used three different scenarios. The first scenario finds a counterexample even when a Max-resolution of 128 is used (i.e., the previous scenario). The second scenario covers the case that no counterexample is generated with a Max-resolution of 128, but there is one with a Max-resolution of 64. In the third scenario, 
there is no counterexample found at all, not even using a Max-resolution of 2.

Table~\ref{tab:results:Matrix} shows the overall time used for the verification of the first scenario depending on the parameters Base- and Max-resolutions ranging from 2 to 64 and from 2 to 128, respectively. The running times from Table~\ref{tab:results:WithRefinement} are included here in the bottom row for facilitating comparisons. 
Note, that also the running times from Table~\ref{tab:results:WithoutRefinement} are included, in the diagonal of Table~\ref{tab:results:Matrix}.
For all combinations with a Max-resolution of 4 and 8, our approach using selective refinement takes slightly more time, but this is in the order of less than a second and, hence, does not really matter. We suspect that this is due to the workflow's overhead for generating the environment model(s), which is small in absolute terms, but compared to these very short verification times, relatively large.
For larger Max-resolutions, this overhead can be neglected.
The shortest verification run that leads to the counterexample with the longest path -- all other counterexamples are considered spurious since they are only due to the structural abstraction -- can be achieved with Max-resolution 64 and Base-resolution 4. Using a Max-resolution of 128 does not make a big difference in terms of verification time. It checks immediately whether the counterexample still exists in the higher-resolution representation.


We also model-checked a second scenario, which is a slightly adapted version of the running example, in order to get results for a {\em near miss} situation.
In this second scenario, we slightly increased the distance between the robot arm and the table by (virtually) expanding the height of the mounting plate slightly. 
We intentionally defined the adapted height in such a way that
a verification with resolution 64 results in a counterexample, while a verification with resolution 128 passes. 
The results are given in Table~\ref{tab:results:Near}. Here a large time difference between the results of Max-resolution 64 and 128 is shown. Since a collision in this scenario is only nearly missed, the verification run with Max-resolution 64 found a counterexample quickly and this run was short, since with this resolution it looks like there is a collision.
With the higher Max-resolution of 128, however, this collision cannot be seen anymore, so that there is no corresponding counterexample, and 
the verification cannot stop early.

We were interested in whether this means that only with real counterexamples our new approach is that efficient as shown in Table~\ref{tab:results:Matrix}.
Hence, we defined a related third scenario, where 
the obstacles are out of the robots reach and, therefore, 
a collision is obviously avoided.
Table~\ref{tab:results:Safe} shows the results for this scenario, and it shows comparable running times to Table~\ref{tab:results:Matrix}.
That is, only in the near-miss situation the running times with Max-resolution 64 were larger.


\fussy
Across all scenarios, the verification workflow with Max-resolution $x$ outperforms the verification without structural abstraction and resolution $x$ while finding the same counterexample in terms of robot position. The Base-resolution only influences the time advantage gained by using our approach.    
\sloppy

The data source as well as the source code of the approach presented here are available in an executable artefact on \url{https://zenodo.org/record/7622703}.

\section{Related Work}



A bounding volume hierarchy is a hierarchy that arranges bounding volumes of objects into a tree structure~\cite[Chapter 6]{TMKS}. It is often used for collision detection, e.g., in game engines.
One way to model a voxel grid as a bounding volume hierarchy is using an octree. 
A node of an octree has exactly eight children or no children at all. Each leaf node of the octree represents a voxel of the voxel grid.  
All other nodes of the octree represent what we call abstract voxels of varying sizes depending on the node's level in the octree. 
However, modeling the whole octree would lead to a very large environment model. That is why our workflow uses environment representations that include only distinct parts of the octree and, hence, the bounding volume hierarchy of the voxel grid, depending on the selective refinements introduced. 

Multi-level voxels models~\cite{youngGPUacceleratedGenerationRendering2018,ghadaiDirect3DPrinting2021} may be an alternative to our approach of representing different resolutions. It uses two types of voxels -- coarse and fine resolution voxels. The coarse resolution voxels form the base voxel grid. At object boundaries, a coarse resolution voxel is subdivided into fine resolution voxels. This enables a more detailed approximation of objects with a lower overall number of voxels as compared to using fine resolution voxels throughout the whole model. Our approach also refines course voxels into finer-grained ones. However, it selectively refines voxels deemed important for a particular verification task, and it does so on several levels of granularity. 

Babić and Hu~\cite{babic2007} introduced structural abstraction of software by providing an automatic abstraction-checking-refinement framework. It also follows the general approach of CEGAR of generating a coarse initial abstraction and refining it based on counterexamples. The framework uses ``the natural function-level abstraction boundaries present in software''\cite{babic2007} to omit details about function behavior. 
Initially, their approach treats the effects of function calls as unconstrained variables. Later constraints are added based on the counterexamples. 
In essence, they use structural information gathered by analyzing the software to guide behavioral abstractions.
Our approach differs since it does not abstract the behavior of a system. Instead, it abstracts structure of the environment the system is embedded into. 

For the verification of hybrid systems based on the CEGAR approach, Clarke et al.~\cite{clarke:2003_2, Clarke:Hybrid:2003} use both structural and behavioral abstractions.
In their example, the width of a street is represented, where a car is not allowed to drive over the borders of the street, neither to the left nor to the right. 
In contrast to this example, the formulas in our example are much more numerous and complex, hence we automatically create them from the voxel representation. In particular, our focus is enterely on the structural abstractions possible here and how to translate them from the voxel representation.

Fishwick and Lee~\cite{fishwick1996two} group the behavior of different (physical) entities in a structure, which is also abstracted in their approach.
One state in their finite state automata (FSA) represents a high-level state of the system. The system's behavior in a specific state is then comprised of the behavior of the individual entities in this state. In contrast, we neither use abstract states nor abstract behavior in our approach.

Yang and Chen~\cite{yang2021} use artificial intelligence (AI) based shape abstraction to map point clouds into a representation consisting only of cuboids. However, inferring that a concrete model satisfies a property when the abstract model satisfies it relies on over-approximation --  in our case, this means that the whole ``real'' object is contained in its abstract representation. Using their method for abstracting the environment does not guarantee over-approximation, however. Therefore, their approach cannot be used as the basis for generating the voxel grids that we need for our approach of selective refinement.

Henzinger et al.~\cite{DBLP:conf/popl/HenzingerJMS02}
introduced
\emph{lazy abstraction}, which
continuously
builds and refines a single abstract model on demand,
driven by the model checker, so that different parts of the
model may exhibit different degrees of precision,
just
enough to verify the desired property.
Our approach uses a similar principle, but it works on structural models as opposed to behavioral models
as implemented in C programs.

Rathmair~et~al.~\cite{rathmair2021}
proposed a verification workflow for robot applications that the one here builds upon.
To overcome the hassles that come with using a high-resolution voxel grid, Rathmair~et~al.\ divide the environment into multiple objects, e.g., table and blue box. Each voxel grid has a different resolution, depending on its size and the details of the object required for the verification. For example, a voxel grid of resolution 16 covering only the blue box has significantly smaller voxels and, therefore, provides more box details than a voxel grid with the same resolution covering the whole environment. 
Our new approach is different because the level of detail is not homogeneous over a whole object. Instead, the model may provide more details of certain parts of an object when they are relevant, and the selective refinement automatically determines where more details are needed.


\section{Discussion}

There is a tradeoff between the number of model-checker runs for refinements and the time needed for each model-checker run. Generally, a model-checker run with a lower resolution takes less time than one with a higher resolution. However, workflow executions with lower Base-resolution need more refinements and, therefore, more individual model-checker runs. Based on our analysis of this tradeoff, determining a value for Base-resolution upfront can be done by educated guessing and, later, through experience.

Although our new verification workflow has shown its potential to strongly outperform verification without refinements for our example, we see room for further improvement. 
Detailed analysis of the individual refinements made while verifying our example showed that generated counterexamples tend to have the same length. That is, the same robot position as in the verification run before causes a counterexample even though a refinement was made. The reason is that a counterexample only highlights one voxel that causes a violation, although, at each robot position, more than one voxel may cause a violation. This may lead to situations where the model-checker finds a violation, a refinement is done, and the next verification step finds a violation at the same robot position. 
We saw this phenomenon when our approach tries to overcome the counterexample at length $21$,
when using a resolution of $32$ without selective refinement.
Manually analyzing this exact robot position revealed that up to $6$ voxels -- dependent on the Base-resolution -- violate the property. However, the workflow refines the voxels one by one, so that
up to $5$ additional model checker runs may be needed. 
Static analysis of the exact situation where the counterexample occurs, and not only refining the voxel given in the counterexample could reduce the number of model-checker runs needed and, hence, the verification time. 

Our proposed workflow does not check the outcome of the refinement of a voxel
for the following situation. 
A SOLID voxel proposed for refinement may only be composed of SOLID higher-resolution voxels, i.e., voxels with the subsequent higher resolution. In such a situation, it is unavoidable that the model-checker detects a collision in its next run. Refining the higher-resolution voxels at the same time,
even over several refinement levels, 
could skip such model-checker runs. In our robot application, such a situation has been encountered multiple times. With a Base-resolution of $4$, already the first refinement has this structure. Introducing multiple refinements in such a situation could have skipped several model-checker runs and, hence, reduced the overall verification time.
There can even be the extreme case that a SOLID voxel proposed for refinement is entirely composed of SOLID voxels even at the Max-resolution. 
In this case, the introduction of new refinements can be skipped since the model-checker is guaranteed to find a counterexample, anyway,
leading to reduced verification time.



Although the structural abstraction proposed in this paper is done on voxel grids only, the overall approach is not limited to them. However, it is necessary to have some representation of the relationship between structural elements on one level of abstraction to elements on another level of abstraction. In the case of voxel grids, the indexes of a voxel allow computing the indexes of the lower-resolution voxel that it is part of and the indexes of all higher-resolution voxels that are abstracted by it. In general, it may be necessary to represent this relationship explicitly.

Finally, since the voxel grid provided to the workflow is already an abstraction of the real environment, it
would have to be made sure
that it is generated in such a way that this abstraction is an ``over-approximation'' of the real environment. Unless this is ensured, however,
the workflow
cannot guarantee that a collision in the real environment is detected.
However, this issue is out of the scope of the workflow, and it arises also when directly performing model-checking without abstractions and refinements, of course.

\section{Conclusion}


In this paper, we propose a new approach to formal verification of safety-critical robot applications, using structural abstraction of environment models and their selective refinement.
While this is inspired by CEGAR, which abstracts and refines behavioral models, our main contribution is to show that major improvements of the efficiency and, hence, applicability of model-checking are feasible even without such changes in a behavioral model.

\bibliographystyle{IEEEtran}
\bibliography{IEEEabrv,my_bib}

\end{document}